# Logical methods of object recognition on satellite images using spatial constraints

R.K. Fedorov

*Institute of System Dynamics and Control Theory, Siberian Division, Russian Academy of Sciences, Irkutsk, Russia*

fedorov@icc.ru

## Abstract

*To solve the problems of urban object recognition a logical method using its structural specifications is proposed. The main idea of the method is to perform the object recognition as a logical inference on the set of rules, describing the correct object shape.*

## 1. Introduction

The problem of urban objects recognition on satellite images is rather complicated. The characteristic features of satellite images are noisiness, blur and low resolution as compared with aerial photo. A satellite image contains a lot of objects to be recognized, which can overlap and can have complex textures (e.g. tile roofing). Textures of adjacent objects can coincide, which results in loss of parts of contours between them. Moreover different object parts can be displayed differently because of illuminations (e.g. roof slopes). As a result, the objects, on the one hand, can be represented by inhomogeneous regions, and, on the other hand, some object regions can be merged.

This article considers processing of grayscale satellite images in visual spectrum. Although grayscale satellite images have less information value, they are cheaper and there're favorable conditions to obtain them without clouds. According to the above mentioned features of satellite images we can state that the methods of their analysis should be able to use some partial object contour information represented by gradients of brightness. Let's consider some local image area containing a straight-line part of an object contour. Usually image brightness varies smoothly along the line orthogonal to the object contour, and it is required to find the exact contour position along the line. There exists a lot of approaches to the problem of contour position detection. For example some methods place contour points on the inflection of brightness function, some methods place contour points on position of the maximum of the brightness function gradient, etc. All of these methods use a limited set of information, considering often only a local area of the image. Therefore their results are often differ from true contours. In Figure 1 (b) the result of processing satellite image fragment by the J. Canny algorithm [1] is shown.

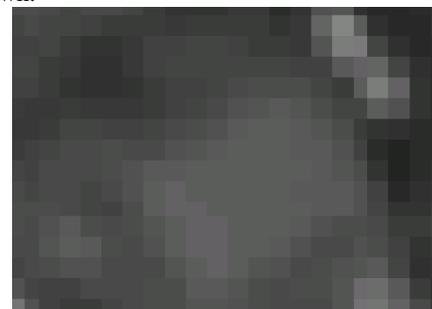
(a)

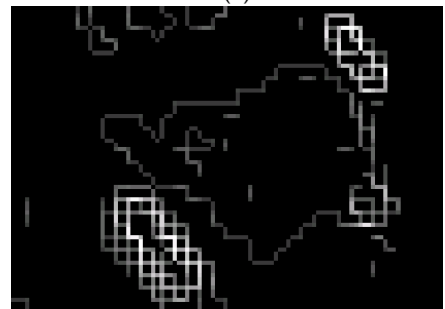
(b)

**Figure 1: (a) source satellite image,
(b) contours obtained by the J. Canny algorithm**

A building is located in the center of the image. The building contour is represented by rather smooth brightness changes. Using the local methods leads to obtaining instead of a single contour segment a set of small contour parts of different directions because of the image noise, inaccuracy of border position detection, and so on. During object recognition the set of contour parts has to be matched against an expected straight-line segment of object contour. Moreover there can exist brightness jumps which aren't part of recognition objects, e.g. ridges of tiles. All of these factors leads to generation of the large set of possible contours, most part of which are wrong. Matching objects to the set is a complex task.

## 2. Main idea

To solve the problems under consideration a logical method of recognition of urban objects using its structural specifications is proposed. The main idea of the method is to perform the object recognition as matching the straight-line segments of the object contour to the brightness jumps of the image. Although many segments can be detected on an image fragment having brightness jump, a logical inference on the set of rules, describing the correct object shape, can be used to limit the set of lines under consideration. Using relative position information of object straight-lines allows to find contours not well defined or defined partially.

## 3. Relative position information

Let's introduce the following notation:
$$f : G \to V, G \subseteq R^2, V \subseteq R^m$$ - raster image.

$G \subseteq R^2$ - set of raster image points.

$V \subseteq R^m$ - set of pixel values (grayscale, RGB, CMYK etc.).

One of the ways to substantially reduce the size of the set of considered straight-lines is to limit their lengths. To calculate the straight-line length we'll use the following auxiliary function, which computes Euclidean distance between two points:
$$len : G^2 \to R$$

Another way to cut down some segments is to pose a constraint on the relative position of object segments, which is defined by computing the angle between two segments with common end point:

$$angle : G^3 \to A, A = [0, 360)$$

Spatial relations between segments can be defined by a set of constraints to these function values.

Let's introduce the function:
$$line : G^2 \to B, B = [0,1]$$

which estimates the strength on the image of the border segment between the two points. Value 1 corresponds to the segment, which is a sharp border between two parts of its neighborhood, value 0 denotes the absence of difference between the parts. In comparison with the detector [2] the given function analyses some neighborhood of a straight-line segment between any two points of the image. Using the function line we can choose between the segments, which are compatible with the position of the whole object and the separate parts of its contour.

## 4. Implementation of the line function

One can distinguish two types of contours on the image:
1) the lines, separating two different areas;
2) the lines, separating two similar areas, but having brightness change along the line itself (boundary line).

It is possible to suggest various estimations of straight-line segment strength. We'll use a heuristic method, which allows to extract contours of the first type. The method is based upon analysis of two rectangles along the segment (Figure 2). The width of rectangles is defined by the method parameter.

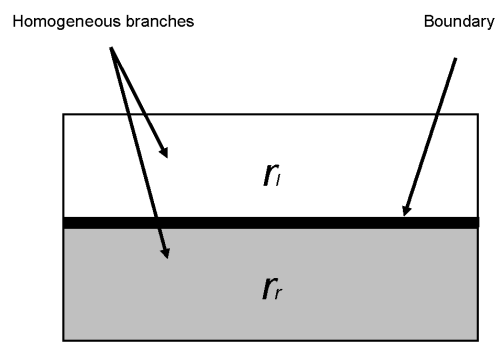

**Figure 2: Analysis area of the line function**

For each rectangle a histogram on $n$ equal intervals is computed. Let's denote by $w_i^r$ the count of pixels from the brightness interval $i$ in the rectangle $r$ The line strength function is defined as:

$$line(p_1, p_2) = 1 - \frac{\sum_1^n \min(w_i^1, w_i^2)}{\sum_1^n \max(w_i^1, w_i^2)}$$

## 5. The procedure of matching a recognition object

Let's define the set of functions of point coordinates computed on the image:

$$Sp = \{line, angle, len\}$$

Let's denote:

$SC = (G, V, f, Sp)$ - a spatial model of processing of image objects.

The procedure of object search within the spatial model can be built as a search of a set of straight-line segments $obj = \{<p_1^1, p_1^2>, ..., <p_k^1, p_k^2>\}$ which satisfy some spatial constraints with the maximum general estimation $w = h(b_1, ..., b_k)$, where $b_i = line(p_i^1, p_i^2)$. The general estimation $h(b_1, ..., b_k)$ can be computed as a linear convolution $h(b_1, ..., b_k) = \sum_{i=1}^k w_i b_i$. The convolution factors can be considered as a measure of importance of the corresponding fragments of the object contour. Therefore this general estimation should be set separately for each object class.

Let's consider the basic properties which the description of object forms with use of the introduced spatial relations should have. We will notice that recognition objects can have various forms (an example – forms of buildings). Spatial relations are set by a set of constraints to values of the spatial functions. And, the concrete recognition object can be defined by conjunction of these constraints. However the real building can have or not have, for example, attach. Therefore disjunction using in the form description increases an expressiveness of the description language [3]. We will note possible recursive definitions of object properties. In particular we can define a building wall recursively as sequence of parts among which there can be a piece with a low estimation. Such definition will help to find contours of walls with ruptures.

The most suitable mechanism of the object form description is descriptive constraint logic programming (CLP). Constraint logic programming is closely connected with traditional logic programming. The majority of constraint logic programming systems represent the interpreter of Prolog with the check of a certain class of constraints built in the mechanism. We will introduce the predicates corresponding to the considered spatial functions on images. These predicates can lead to adding constraints in a base of constraints and its check on compatibility.

1. line (p1, p2, w) - calculates factor w for points p1 and p2.
2. angle (p1, p2, p3, res) calculates degrees of an angle between straight-line segments (p1, p2) and (p2, p3).
3. len (p1, p2, res) – distance between two points p1, p2.

There are many implementations of CLP interpreter. However all of them support only a certain set of constraints. So, the constraint as length (p1, p2, L) &L <10 at known p1 isn't realized without the consideration of all possible values p2 from domain (image) and distance calculation.

## 6. Approbation

For check of the offered method the search of objects in Borland Delphi is realized. The search mechanism engine containing the following object form description expressed by Prolog is realized:

*house(p1,p2,p3,p4) :- line(p1,p2,b1), b1>0.8, angle(p1, p2, p3, l1), l1=90, line(p2,p3,b2), b2>0.8, angle(p2, p3,p4, l2), l2=90, line(p3,p4,b3), b3>0.8, line(p4,p1,b4).*

This description corresponds to rectangular objects.

The experiment was spent on a set of images, including Irkutsk city image fragments with the resolution of 0.7 meters on pixel. The results of experiment are shown on Figures 3,4,5.

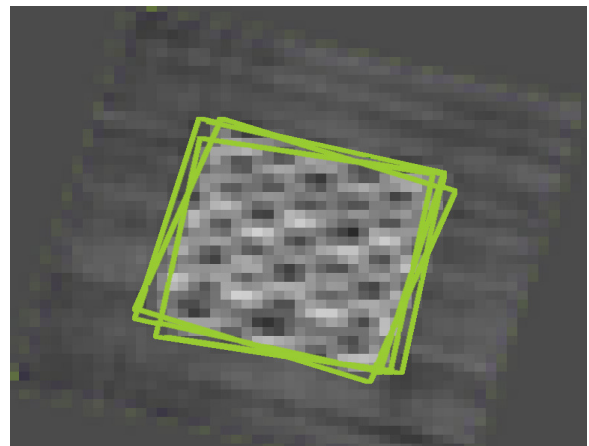

**Figure 3: The result of processing a test image**

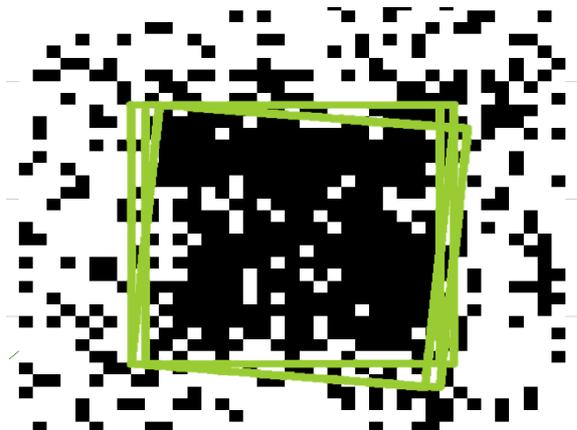
**Figure 4: The result of processing a test image**

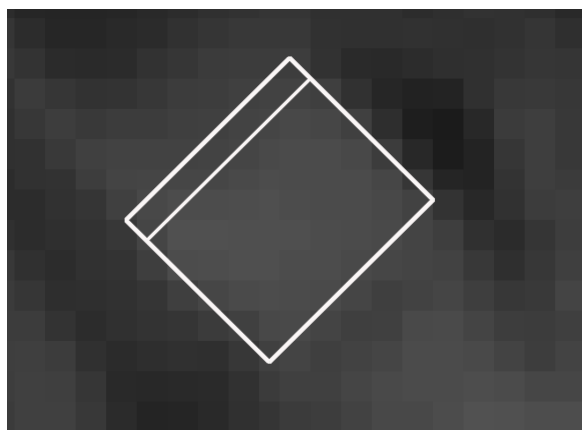
**Figure 5: The result of processing an Irkutsk city image fragment**

## 7. The conclusion

The approbation results confirm efficiency of the method: use of the additional structural information during segmentation and recognition of objects on the image. A lack is computing complexity. Further it is planed to develop the offered method – to develop convenient language of the object form description and to develop the interpreter of language which would be effectively on time and to quality to find recognized objects.